\documentclass[]{template}

\usepackage{wrapfig}
\usepackage{tabularx}
\usepackage{textcomp}
\usepackage{stfloats}
\usepackage{url}
\usepackage{verbatim}
\usepackage{graphicx}
\usepackage{titlesec}
\usepackage{tocloft}
\usepackage{adjustbox}
\usepackage{multirow}
\usepackage{pifont}
\usepackage{tikz}
\usepackage{comment}
\usepackage{amsmath,amssymb} 
\usepackage{colortbl}  
\usepackage{color}
\usepackage{booktabs} 
\usepackage{hyperref}
\usepackage{graphicx}   
\usepackage{subcaption} 
\usepackage{booktabs}
\usepackage{amsmath} %
\usepackage{multirow} %
\usepackage{booktabs} %
\usepackage{subcaption} %
\usepackage{graphicx}   %
\usepackage{wrapfig}    %
\usepackage{caption}    %
\usepackage{attachfile2}
\usepackage{marvosym}
\newcommand{\emailsymbol}{\mbox{\scriptsize\Letter}}

\NewDocumentCommand{\audiobutton}{O{\faMusic} O{Click to play} O{blue!70} m m}{%
   \textattachfile[color=1 1 1, mimetype=audio/mp3]{#4}{%
     \colorbox{#3}{\color{white}#1 ~~#2 \ \raisebox{-0.25em}{\includegraphics[height=1.2em]{#5}}}%
   }%
}
\RequirePackage{xspace}
\makeatletter
\DeclareRobustCommand\onedot{\futurelet\@let@token\@onedot}
\def\@onedot{\ifx\@let@token.\else.\null\fi\xspace}

\makeatother

\usepackage{makecell}
\definecolor{Gray}{gray}{0.95}

\definecolor{adptorange}{RGB}{248, 205, 172}
\definecolor{cmpblue}{RGB}{189, 215, 238}

\definecolor{our_red}{RGB}{232,157,160}
\definecolor{our_blue}{RGB}{136,206,230}
\definecolor{our_orange}{RGB}{246,200,168}
\definecolor{our_green}{RGB}{178,211,164}

\definecolor{attn_code0}{RGB}{247,215,200}
\definecolor{attn_code1}{RGB}{238,169,139}
\definecolor{mlp_code0}{RGB}{204,201,221}
\definecolor{mlp_code1}{RGB}{102,95,153}

\definecolor{token_blue}{RGB}{84, 120, 140}

\definecolor{myMagenta}{rgb}{0.9,0,0.4}

\usepackage{pifont}       
\usepackage{fontawesome}
\usepackage{xspace}

\usepackage{float}

\newcolumntype{x}[1]{>{\centering\arraybackslash}p{#1pt}}
\newcolumntype{y}[1]{>{\raggedright\arraybackslash}p{#1pt}}
\newcolumntype{z}[1]{>{\raggedleft\arraybackslash}p{#1pt}}

\usepackage{colortbl}
\usepackage{xcolor}
\usepackage{wrapfig}

\renewcommand{\paragraph}[1]{\vspace{1.25mm}\noindent\textbf{#1}}

\usepackage{algorithm}
\usepackage{listings}

\definecolor{codeblue}{rgb}{0.25, 0.5, 0.5}
\definecolor{codekw}{rgb}{0.35, 0.35, 0.75}
\lstdefinestyle{Pytorch}{
    language = Python,
    backgroundcolor = \color{white},
    basicstyle = \fontsize{9pt}{8pt}\selectfont\ttfamily\bfseries,
    columns = fullflexible,
    aboveskip=1pt,
    belowskip=1pt,
    breaklines = true,
    captionpos = b,
    commentstyle = \color{codeblue},
    keywordstyle = \color{codekw},
}

\definecolor{green}{HTML}{009000}
\definecolor{red}{HTML}{ea4335}

\title{CoinVE-200K: A Large-Scale High-Quality Dataset for Compositional Instruction-Guided Video Editing}

\author{Smart Creation Platform Department, Online Video BU, Tencent}

\newauthor{Fuchen Long, Cong Wang, Zitao Gao, Wenhao Zhong, Yu Cheng, Xiaolu Hou, Yan Li, Xiao Cao, Xinlong Sun$^{\dag}$, Xi Chen$^{\emailsymbol}$, Yu Liu}
\affiliation[\dag]{\scriptsize{Project Leader}}
\affiliation[\emailsymbol]{\scriptsize{Corresponding Author}}

\abstract{
The quality and diversity of instruction-based video editing datasets are steadily improving, yet existing datasets mainly focus on single editing operations and fall short in supporting compositional instruction-guided video editing.
Particularly, multiple editing intents need to be jointly understood and faithfully executed within the same video.
To address this issue, we introduce \textbf{CoinVE-200K}, a large-scale, high-quality dataset for \textbf{Co}mpositional \textbf{In}struction-Guided \textbf{V}ideo \textbf{E}diting.
The proposed dataset contains \textbf{1080p} video-editing pairs of up to \textbf{201} frames covering diverse compositional editing scenarios, where each sample involves $2$ to $5$ atomic editing operations.
Editing instructions span multiple target subjects, including humans, objects, and backgrounds, covering a broad range of edit types such as addition, removal, modification, and stylization.
All samples are constructed through a carefully designed data generation and quality filtering pipeline to ensure instruction faithfulness, visual quality, temporal consistency, and compositional diversity. Compared with existing instruction-based video editing datasets, CoinVE-200K places a stronger emphasis on multi-intent composition, region-aware editing, and complex interactions among different editing operations.
Besides, to provide a unified evaluation protocol for this challenging setting, we introduce \textbf{CoinVE-Bench}, a dedicated benchmark for compositional-instruction video editing, covering diverse combinations of editing subjects, operation types, and instruction complexities.
We further present \textbf{CoinVE-Edit}, a 22B compositional video editing model built upon Wan2.1-T2V-14B and Qwen3VL-8B.
CoinVE-Edit disentangles region-aware attention for different editing instructions, enabling precise multi-region editing while preserving irrelevant content and maintaining temporal coherence. Extensive experiments on CoinVE-Bench demonstrate that CoinVE-Edit achieves strong performance in instruction following, compositional editing accuracy, visual quality, and temporal consistency, providing a powerful baseline for future research on compositional instruction-guided video editing.
}

\date{August, 2026}
\webservice{\url{https://coinve200k.github.io}}
\data{\url{https://huggingface.co/datasets/FireCRT/CoinVE-200K}}

\begin{document}
\makeatletter
\let\CT@arc@\relax
\let\CT@drsc@\relax
\makeatother
\thispagestyle{firstheader}
\maketitle
\pagestyle{empty}

\section{Introduction}

The recent success of instruction-guided image editing models, including FLUX-Kontext\citep{flux}, Qwen-Image-Edit\citep{qwenimage}, and Nano Banana\citep{comanici2025gemini}, highlights their strong capability in understanding user intent and executing high-quality visual modifications. This success is closely tied to the rapid development of large-scale, diverse, and high-quality datasets for instruction-based image editing\citep{nhredit,gpt-image-edit,picobanana}.
Meanwhile, instruction-guided video editing has also attracted increasing attention due to its broad applications in content creation, film production, advertising, and interactive visual storytelling. Compared with image editing, video editing requires not only accurate instruction following and spatially precise modifications, but also strong temporal consistency across frames, making the construction of high-quality video editing datasets substantially more challenging.

Existing instruction-guided video editing datasets have made important progress in scaling up data and covering diverse editing categories. However, most of them mainly focus on single editing operations, such as changing a background, removing an object or modifying a local region. While these datasets are useful for training models to perform isolated edits, they are insufficient for practical editing scenarios, where users often provide multiple editing intentions within a single request. For example, a user may ask a model to replace a person, remove an object, and stylize the background simultaneously. Such compositional editing requires the model to jointly understand multiple instructions, identify their corresponding spatial-temporal editing regions, execute each operation faithfully, and preserve irrelevant content. This setting is significantly more difficult than single-operation editing, as different editing intents may interact with each other, leading to instruction interference, unintended region modification, and temporal inconsistency.

\begin{figure}[t]
\centering
\includegraphics[width=0.99\linewidth]{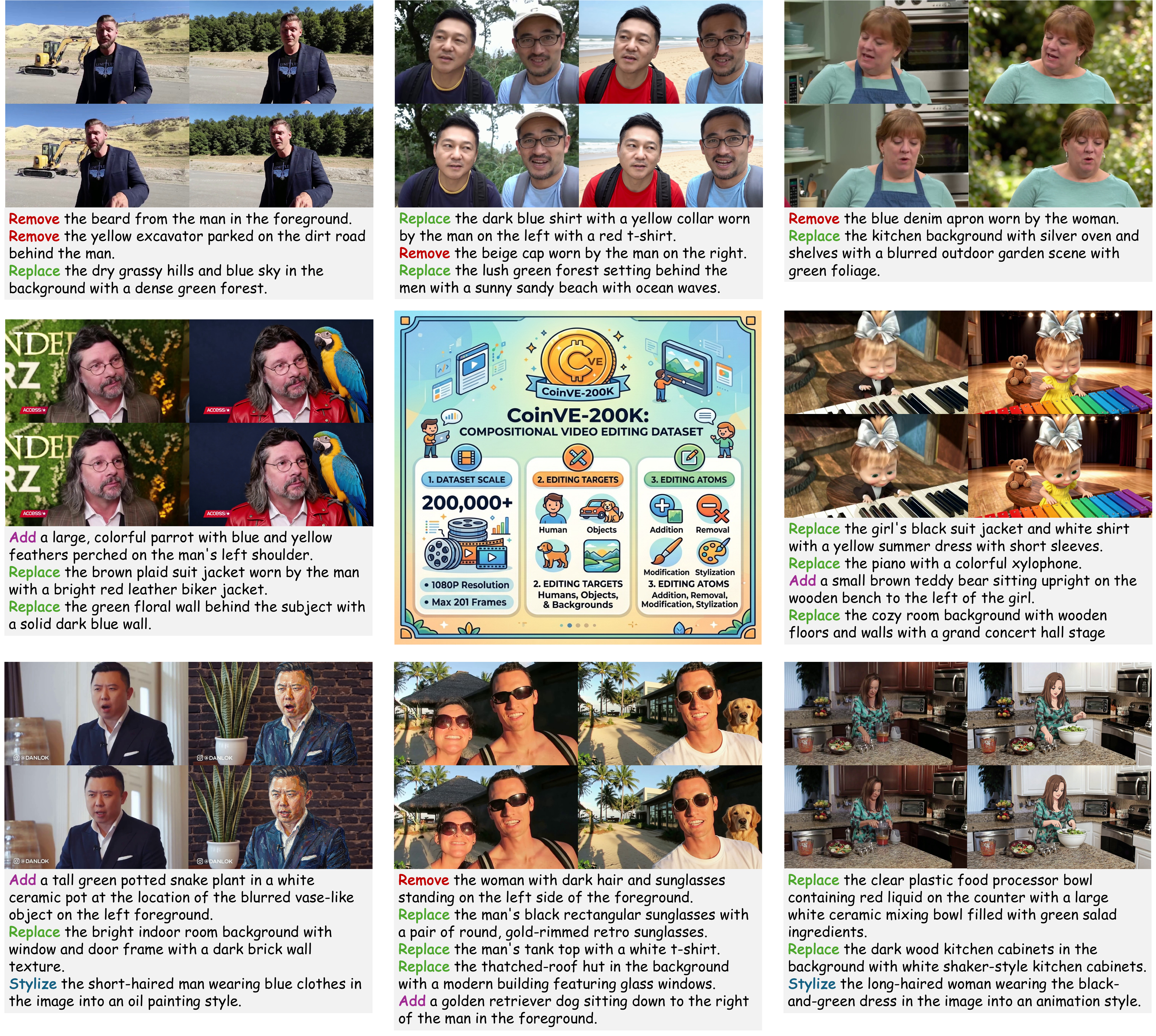}
\caption{Demonstration of compositional instruction-guided video editing cases from our proposed CoinVE-200K.}
\vspace{-0.15in}
\label{fig:dataset_teaser}
\end{figure}

We refer to this challenging setting as \textit{Compositional Instruction-Guided Video Editing}, where each editing request consists of multiple atomic operations that should be executed coherently within the same video. Despite its practical importance, this problem remains under-explored. Existing instruction-guided video editing datasets~\citep{openve,ditto,reco} usually suffer from three limitations when applied to compositional editing. First, their editing instructions are often short and correspond to only one dominant editing intent, making it difficult for models to learn multi-intent reasoning. Second, the editing targets are usually limited to a single subject or region, while real-world compositional editing may involve humans, objects, and backgrounds simultaneously. Third, existing datasets lack systematic annotations or data construction strategies for combining different edit types, such as addition, removal, modification, and stylization. Consequently, models trained on these datasets often struggle to perform complex multi-operation editing and tend to modify irrelevant regions or ignore part of the given instructions.

To address these limitations, we introduce \textbf{CoinVE-200K}, a large-scale, high-quality dataset for \textbf{Co}mpositional \textbf{In}struction-Guided \textbf{V}ideo \textbf{E}diting. The proposed dataset contains 200K video-editing pairs, each in \textbf{1080p} resolution with up to \textbf{201} frames, covering diverse compositional editing scenarios.
Each sample involves $2$ to $5$ atomic editing operations, encouraging models to learn how to decompose, associate, and execute multiple editing intents within a unified generation process.
As shown in Figure~\ref{fig:dataset_teaser}, the editing instructions span multiple target subjects, including humans, objects, and backgrounds, and cover a broad range of edit types, including addition, removal, modification, and stylization. Compared with existing instruction-based video editing datasets, CoinVE-200K places stronger emphasis on multi-intent composition, region-aware editing, and complex interactions.
The construction of CoinVE-200K is supported by a carefully designed data generation and quality filtering pipeline. 
Given source videos, we first analyze their visual content and identify editable subjects, including foreground humans, salient objects, and background regions. Then, we generate compositional editing instructions by combining multiple atomic operations under predefined editing taxonomies. These instructions are further used to produce edited videos while maintaining spatial plausibility and temporal coherence. To ensure data quality, we adopt a multi-stage quality filtering strategy that evaluates generated video-editing pairs from several perspectives, including instruction faithfulness, visual quality, temporal consistency, subject preservation, and compositional correctness. This pipeline enables CoinVE-200K to provide high-quality supervision for complex compositional video editing.

Based on CoinVE-200K, we further present \textbf{CoinVE-Edit}, a 22B compositional video editing model built upon Wan2.1-T2V-14B~\citep{wan2025wan} and Qwen3-VL-8B-Instruct~\citep{qwen3vl}. CoinVE-Edit is designed to effectively handle multi-instruction editing requests by disentangling region-aware attention for different editing instructions. Specifically, the model first leverages the multimodal understanding capability of Qwen3-VL to interpret both the input video and the compositional editing instruction. Then, the video generation backbone based on Wan2.1 performs instruction-guided video transformation. To mitigate interference among different editing operations, CoinVE-Edit introduces a region-aware attention disentanglement mechanism, where different editing instructions are associated with their corresponding spatial-temporal regions and controlled independently during generation. This design enables precise multi-region editing while preserving irrelevant content and maintaining temporal coherence.

In addition to the dataset and model, we propose \textbf{CoinVE-Bench}, a dedicated benchmark for compositional instruction-guided video editing. Existing video editing benchmarks mainly evaluate single-operation editing and are insufficient for measuring compositional editing ability. CoinVE-Bench covers diverse combinations of editing subjects, operation types, and instruction complexities. It evaluates model performance along multiple dimensions, including instruction following, compositional editing accuracy, region-level editing precision, irrelevant content preservation, visual quality, and temporal consistency. 
Technically inspired by recently proposed novel CoVEBench~\citep{covebench} for complex-instruction video editing, CoinVE-Bench combines checklist-based and specialized model evaluation to comprehensively assess video editing quality.

Extensive experiments demonstrate that CoinVE-Edit achieves strong performance on CoinVE-Bench, especially in challenging cases involving multiple editing operations and multiple target regions. Compared with existing instruction-guided video editing models, CoinVE-Edit shows better compositional instruction following, more accurate region-aware editing, improved non-target preservation, and stronger temporal stability. We believe that CoinVE-200K, CoinVE-Bench, and CoinVE-Edit together provide a comprehensive foundation for advancing research on compositional instruction-guided video editing.
In summary, our contributions are:

\begin{itemize}
    \item We introduce \textbf{CoinVE-200K}, a large-scale and high-quality dataset for compositional instruction-guided video editing. It contains 200K video-editing pairs, where each sample consists of $2$ to $5$ atomic editing operations. The editing taxonomy covers multiple target subjects, including humans, objects, and backgrounds, and diverse edit types, including addition, removal, modification, and stylization.
    
    \item We present \textbf{CoinVE-Edit}, a 22B compositional video editing model that disentangles region-aware attention for different editing instructions, enabling precise multi-region editing while preserving irrelevant content.
    
    \item We establish \textbf{CoinVE-Bench}, a dedicated benchmark for compositional instructional video editing, evaluating models across instruction following, compositional editing accuracy, region-level precision, non-target preservation, visual quality, and temporal consistency.
\end{itemize}

\section{Related Work}

\subsection{Video Understanding}

Video understanding aims to perceive, describe, and reason about dynamic visual content across both spatial and temporal dimensions. Early studies mainly focus on task-specific video representation learning, such as action recognition~\citep{tran2015learning,wang2016temporal}, temporal localization~\citep{long2019gtan}, video captioning~\citep{zhou2018cap}, and video question answering~\citep{chen2020uniter}. With the rapid development of large language models (LLMs) and multimodal large language models (MLLMs), recent works have shifted toward general-purpose video understanding through multimodal instruction tuning. Representative models, such as Video-LLaMA~\citep{zhang2023videollama}, VideoChat~\citep{li2023videochat}, Video-ChatGPT~\citep{maaz2024videochatgpt}, Video-LLaVA~\citep{lin2024videollava}, and QwenVL series~\citep{bai2023qwenvl,bai2025qwen25vl,qwen3vl}, extend image-language models to video inputs by incorporating temporal visual tokens, video encoders, or video-specific instruction data. Video-LLaMA~\citep{zhang2023videollama} connects video and audio encoders with LLMs for audio-visual instruction following, while VideoChat~\citep{li2023videochat} builds an interactive video-language dialogue system by aligning video foundation models with LLMs. Video-ChatGPT~\citep{maaz2024videochatgpt} further improves detailed video understanding through video feature projection and instruction tuning. Video-LLaVA~\citep{lin2024videollava} learns unified visual representations for both images and videos before language projection, improving cross-modal alignment. More recently, the QwenVL series~\citep{bai2023qwenvl,bai2025qwen25vl,qwen3vl} demonstrates strong multimodal perception, grounding, OCR, and long-context visual reasoning capabilities, making it suitable for extracting high-level semantic conditions from complex video inputs.

For instruction-guided video editing, such video understanding capability is essential for identifying editable subjects, parsing user intentions, and grounding editing operations to corresponding spatial-temporal regions. This becomes more challenging in compositional instruction-guided video editing, where multiple editing intents must be jointly decomposed, grounded, and executed while preserving irrelevant content. Our work follows this direction by leveraging MLLM-based semantic understanding to obtain structured editing conditions and further injecting them into a controllable video editing framework.

\subsection{Video Generation}

Video generation has achieved significant progress with the development of diffusion models and large-scale generative foundation models. Early video diffusion methods extend image diffusion models to the temporal domain by introducing temporal layers, motion modules, or 3D U-Net~\citep{rombach2022sd,singer2023makeavideo,ho2022imagenvideo,long2024vs,2025motionpro,chen2025ourobors,chen2025iccv}. Stable Video Diffusion~\citep{blattmann2023svd} further demonstrates the effectiveness of large-scale image-to-video pretraining and becomes an important foundation for high-quality video synthesis. Subsequent works explore various architectures and training strategies to improve motion realism, temporal consistency, and video quality~\citep{chen2024sateco}, including latent video diffusion models and cascaded video diffusion models.

Recently, Diffusion Transformer (DiT)-based architectures~\citep{opensora,ma2025latte,yang2024cogvideox,kong2024hunyuanvideo} have emerged as a dominant paradigm for scalable video generation, shifting video diffusion models from U-Net-style temporal extensions toward token-based spatial-temporal modeling. Latte~\citep{ma2025latte} first validates this direction by representing videos as latent spatial-temporal tokens and modeling them with transformer blocks. Subsequent large-scale systems further strengthen this paradigm: CogVideoX~\citep{yang2024cogvideox} combines a 3D causal VAE with an expert transformer to improve video compression and text-video alignment, while HunyuanVideo~\citep{kong2024hunyuanvideo} develops a system-level framework with strong text encoders, scalable DiT backbones, and progressive training strategies for high-quality video generation. Meanwhile, LTX-Video~\citep{haocohen2025ltx,haocohen2026ltx2} pushes generation efficiency through highly compressed video latents, and Wan~\citep{wan2025wan} provides a powerful open video foundation model with strong text-to-video and image-to-video capabilities.
These models provide strong generative priors for video editing, but editing requires preserving the input video's identity, layout, and motion while modifying only instruction-relevant regions. Therefore, we build upon a strong video generation backbone and introduce region-aware attention disentanglement to adapt it to compositional instruction-guided video editing.

\subsection{Instruction-Guided Video Editing}

Instruction-guided video editing aims to modify a given video according to natural language instructions. Compared with traditional video editing methods that rely on masks, scribbles, keyframes, or manually specified control signals, instruction-based editing provides a more intuitive and user-friendly interface. Inspired by the success of instruction-guided image editing models such as InstructPix2Pix~\citep{brooks2023instructpix2pix}, MagicBrush~\citep{zhang2023magicbrush}, FLUX-Kontext~\citep{flux}, and Qwen-Image-Edit~\citep{qwenimage}, recent studies have begun to explore natural language-driven video editing. Existing methods usually adopt pretrained video diffusion models as generative backbones and inject editing conditions through additional encoders~\citep{decart2025lucyedit}, cross-attention modules~\citep{tokenflow2024}, adapters~\citep{jiang2025vace}, or feature modulation mechanisms.

A key challenge in instruction-guided video editing is how to preserve the spatial-temporal structure of the input video while faithfully applying modifications. Early methods mainly adapt image diffusion models to videos and enforce temporal consistency through attention control or feature propagation. For example, TokenFlow~\citep{tokenflow2024} propagates diffusion features across frames to obtain temporally consistent edits. With the development of stronger video diffusion and DiT backbones, recent methods increasingly rely on explicit condition injection modules, such as additional encoders, adapters, or multimodal instruction representations. VACE~\citep{jiang2025vace}, for instance, introduces adapter-style conditioning to incorporate diverse video editing signals into the generation process, while recent MLLM-based methods~\citep{lin2026kiwi,zhang2026sama,chen2026vino,pan2026omniweaving} extract high-level instruction semantics and inject them into video generative backbones. These works form a common pipeline of instruction-guided video editing: semantic understanding of the input video and editing instruction, followed by conditional video generation with a pretrained diffusion or DiT backbone.

Recent datasets and models, such as InsViE~\citep{insvie2025}, Senorita~\citep{senorita2025}, Ditto~\citep{ditto}, and OpenVE~\citep{openve}, have advanced data-driven instruction-based video editing with large-scale video-edit pairs. However, they mainly focus on single editing operations, while real-world users often provide compositional instructions involving multiple intents, subjects, and edit types within the same video. Such scenarios require models to decompose instructions, ground each operation to the corresponding spatial-temporal region, coordinate multiple edits without interference, and preserve irrelevant content.
To address this challenge, we introduce \textbf{CoinVE-200K}, a large-scale dataset for compositional instruction-guided video editing, together with \textbf{CoinVE-Bench} for systematic evaluation. We further develop \textbf{CoinVE-Edit}, combining MLLM-based instruction understanding with DiT-based video generation and disentangling region-aware attention for precise multi-region editing.

\begin{figure}[h]
    \centering
    \includegraphics[width=\linewidth]{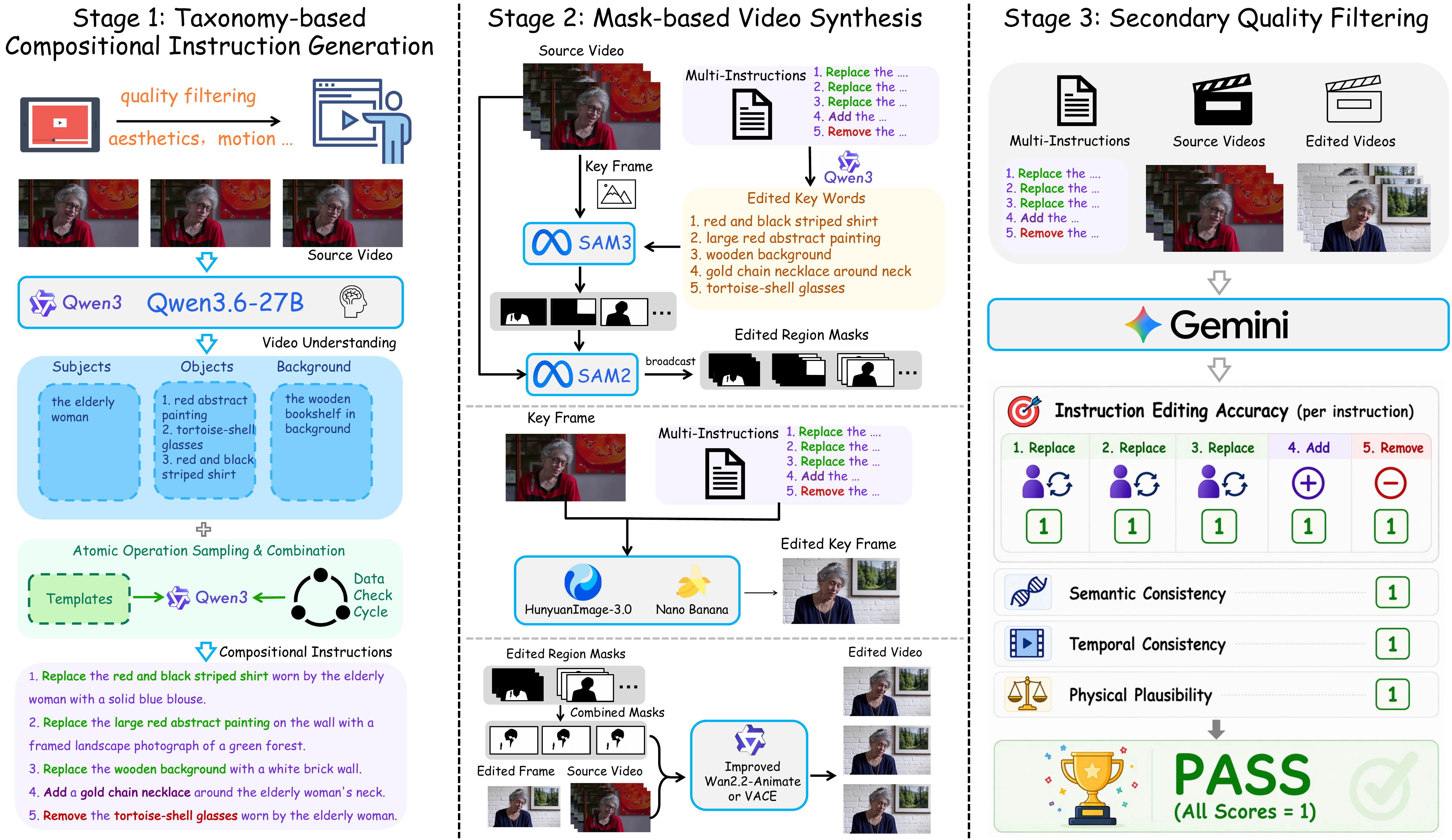}
    \vspace{0.01in}
    \caption{Data construction pipeline of CoinVE-200K.}
    \vspace{-0.01in}
    \label{fig:data_pipeline}
\end{figure}

\section{CoinVE-200K}
Here, we introduce CoinVE-200K, a large-scale high-quality video dataset designed for compositional instruction-guided video editing.
Figure~\ref{fig:data_pipeline} depicts the whole data construction pipeline.
 
\subsection{Video Pre-Processing}
The source video set is collected from the high-quality open-source video dataset, i.e., OpenVid-HD~\citep{nan2025openve}.
Given the whole set, we first filter the video clips based on the frame number and resolution. 
The retained videos are required to contain at least $81$ frames and have a shorter side of at least $1080$ pixels.
Then, we utilize aesthetic scores~\citep{improved-aesthetic-predictor} and optical flow~\citep{raft} to select videos characterized by high aesthetic quality and appropriate motion magnitude.
Finally, we attain $420K$ videos for the subsequent video pair synthesis.

\subsection{Taxonomy-based Compositional Instruction Generation}
To facilitate the instruction generation, the taxonomy of the visual content in the source video should be first parsed. 
As shown in the left part of Figure~\ref{fig:data_pipeline}, we feed the source video into Qwen3.6-27B~\citep{qwen3.6-27b} for video understanding.
For each retained video, Qwen3.6-27B is prompted to summarize its editable visual content from three complementary aspects: \textit{subjects}, \textit{objects}, and \textit{background}. This taxonomy provides a structured description of the source video, including the main actors, salient manipulable entities, and scene context, which serves as the basis for subsequent compositional instruction generation.

Based on the parsed taxonomy, we construct compositional instructions by sampling and combining multiple atomic editing operations. Concretely, we maintain a set of operation templates covering common edit types, such as \textit{replace}, \textit{add}, \textit{remove}, and \textit{stylization}. Qwen3 is then used to instantiate these templates with concrete visual concepts extracted from the source video, yielding semantically grounded atomic instructions. Multiple atomic operations are further composed into a single multi-instruction editing request, so that one sample may simultaneously involve subject appearance modification, object replacement, scene transformation, and attribute addition or deletion. 
To improve the validity and diversity of generated instructions, we additionally apply a data checking cycle to verify that each atomic edit is compatible with the source content and that the combined instruction remains coherent and executable.
Meanwhile, the balance across all types of atomic editing operations is also achieved.

\subsection{Mask-based Video Synthesis}
After obtaining the compositional instructions, we synthesize the edited video in a mask-guided manner, as illustrated in the middle part of Figure~\ref{fig:data_pipeline}. Given a source video and its corresponding multi-instruction prompt, we first sample a representative key frame from the video. Since compositional editing usually targets multiple entities or regions, we use Qwen3.6-27B~\citep{qwen3.6-27b} to extract the edited keywords from the instruction set, i.e., the concrete visual concepts that need to be modified, inserted, or removed. These keywords provide explicit textual anchors for localizing the target regions in the key frame.

We then employ SAM3~\citep{sam3} to segment the relevant regions on the key frame according to the extracted editing keywords. The resulting masks identify the spatial locations associated with each atomic edit, such as a piece of clothing, a wall painting, an accessory, or the background region. To extend these edits from a single frame to the entire video, we further use SAM2~\citep{sam2} to propagate the key-frame masks across adjacent frames, producing temporally aligned edited-region masks for the full video. This step establishes explicit spatial-temporal correspondences for the regions affected by the various instructions.

With the localized masks, we first edit the key frame using strong image editing models, including HunyuanImage-3.0~\citep{hunyuanimage3} and Nano Banana~\citep{gemini2}, conditioned on the multi-instruction prompt. This produces an edited key frame that reflects the desired compositional modifications. 
Afterwards, we combine all propagated region masks into one mask, and further feed it with the edited frame and the original source video into our deliberately fine-tuned Wan2.2-Animate-14B~\citep{wananimate} or VACE~\citep{jiang2025vace} model to synthesize the final edited video. 
Compared with directly editing video through SOTA editing models, this mask-based pipeline provides better control over the edited regions, improves faithfulness to each atomic instruction, and helps preserve temporal consistency across frames.

\subsection{Secondary Quality Filtering}
Although the mask-based synthesis pipeline produces candidate edited videos with diverse compositional modifications, the generated results may still suffer from incomplete instruction execution, temporal inconsistency, or unrealistic visual artifacts. Therefore, as shown in the right part of Figure~\ref{fig:data_pipeline}, we perform a secondary quality filtering stage to retain only high-quality samples for the final dataset.

Specifically, we feed the multi-instructions, source videos, and edited videos into Gemini 2.5 Pro~\citep{gemini2} for automatic quality assessment. The evaluation is conducted from four perspectives. First, we measure \textit{instruction editing accuracy} at the level of each atomic instruction, checking whether every individual operation in the compositional prompt has been successfully executed. This per-instruction verification is particularly important. The sample is only considered valid when all sub-edits are correct. 
Second, we evaluate the overall quality of the edited video from three aspects: semantic consistency, temporal consistency, and physical plausibility. This ensures that the edited result remains coherent with the source content except for the intended changes, stable across frames without flickering or drifting, and physically plausible in geometry, interactions, and scene composition.

We adopt a strict filtering strategy and only keep samples that pass all evaluation items. In practice, this means that each atomic instruction must be marked as successfully executed, while semantic consistency, temporal consistency, and physical plausibility must all receive positive judgments. Such a conservative selection criterion effectively removes low-quality generations and substantially improves the reliability of the final compositional video editing dataset.

\begin{figure}[t]
    \centering
    \includegraphics[width=\linewidth]{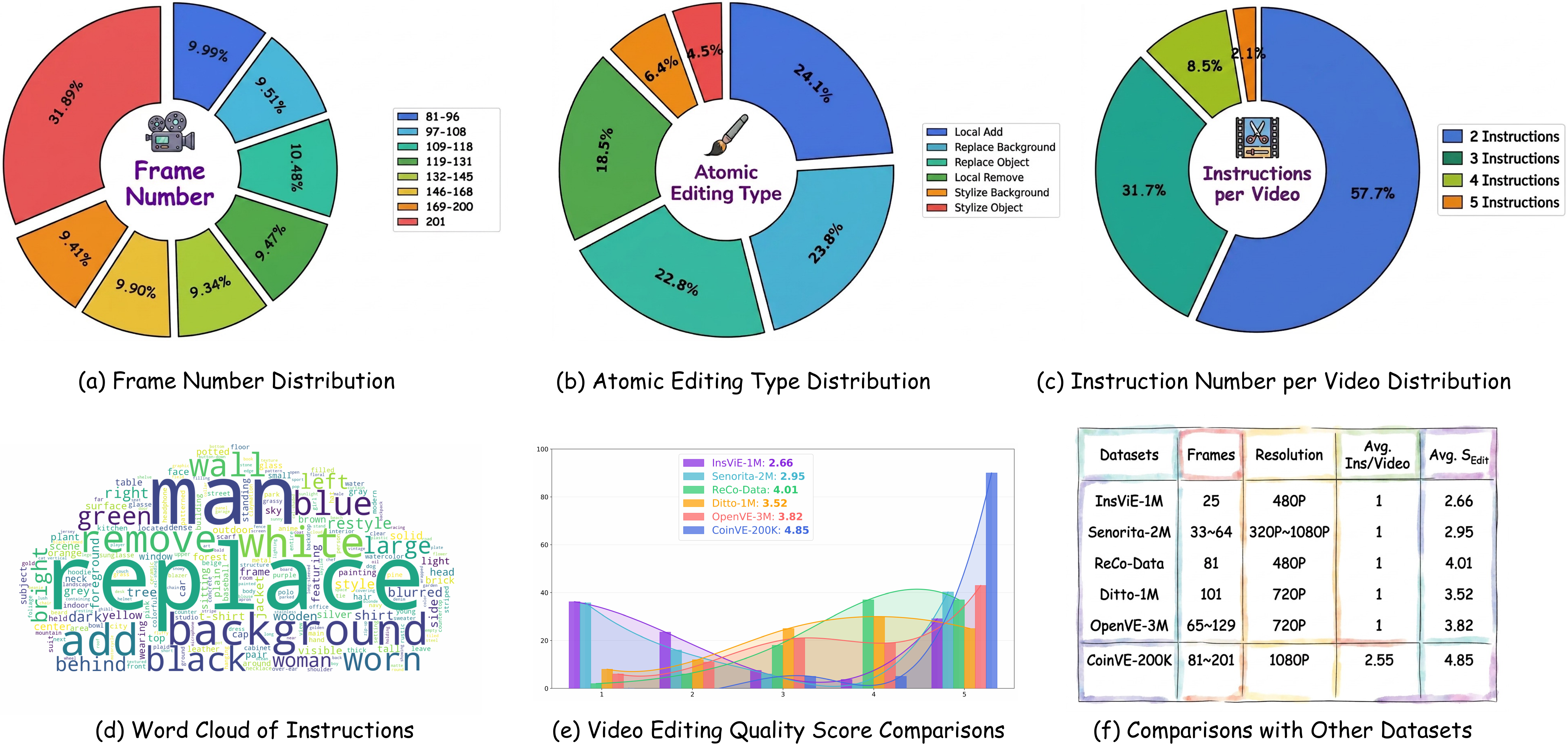}
    \vspace{0.01in}
    \caption{Data statistics of CoinVE-200K.}
    \vspace{-0.01in}
    \label{fig:data_statistics}
\end{figure}

\subsection{Data Statistics}

After the overall data generation pipeline and data quality filtering, we obtain CoinVE-200K, a high-quality compositional-instruction video editing dataset with $200$K samples. As shown in Figure~\ref{fig:data_statistics}, the retained videos contain \textbf{81--201} frames with a resolution of \textbf{1080p}, covering diverse temporal lengths and visual contents. 
CoinVE-200K includes six atomic editing types with a relatively balanced distribution: local addition ($24.1\%$), background replacement ($23.8\%$), object replacement ($22.8\%$), local removal ($18.5\%$), background stylization ($6.4\%$), object stylization ($4.5\%$). Each video is paired with multiple atomic instructions, where $57.7\%$, $31.7\%$, $8.5\%$, and $2.1\%$ of videos contain $2$, $3$, $4$, and $5$ instructions, respectively, resulting in an average of $2.55$ instructions per video. The word cloud further shows that our instructions cover a wide range of objects, attributes, colors, and editing actions, such as ``person'', ``background'', ``add'', and ``remove'', indicating rich linguistic and visual diversity. Compared with existing single-instruction video editing datasets, CoinVE-200K provides richer compositional supervision and achieves a higher average editing quality score of $4.85$, demonstrating both its diversity and reliability for training compositional-instruction video editing models.

\begin{figure}[t]
    \centering
    \includegraphics[width=\linewidth]{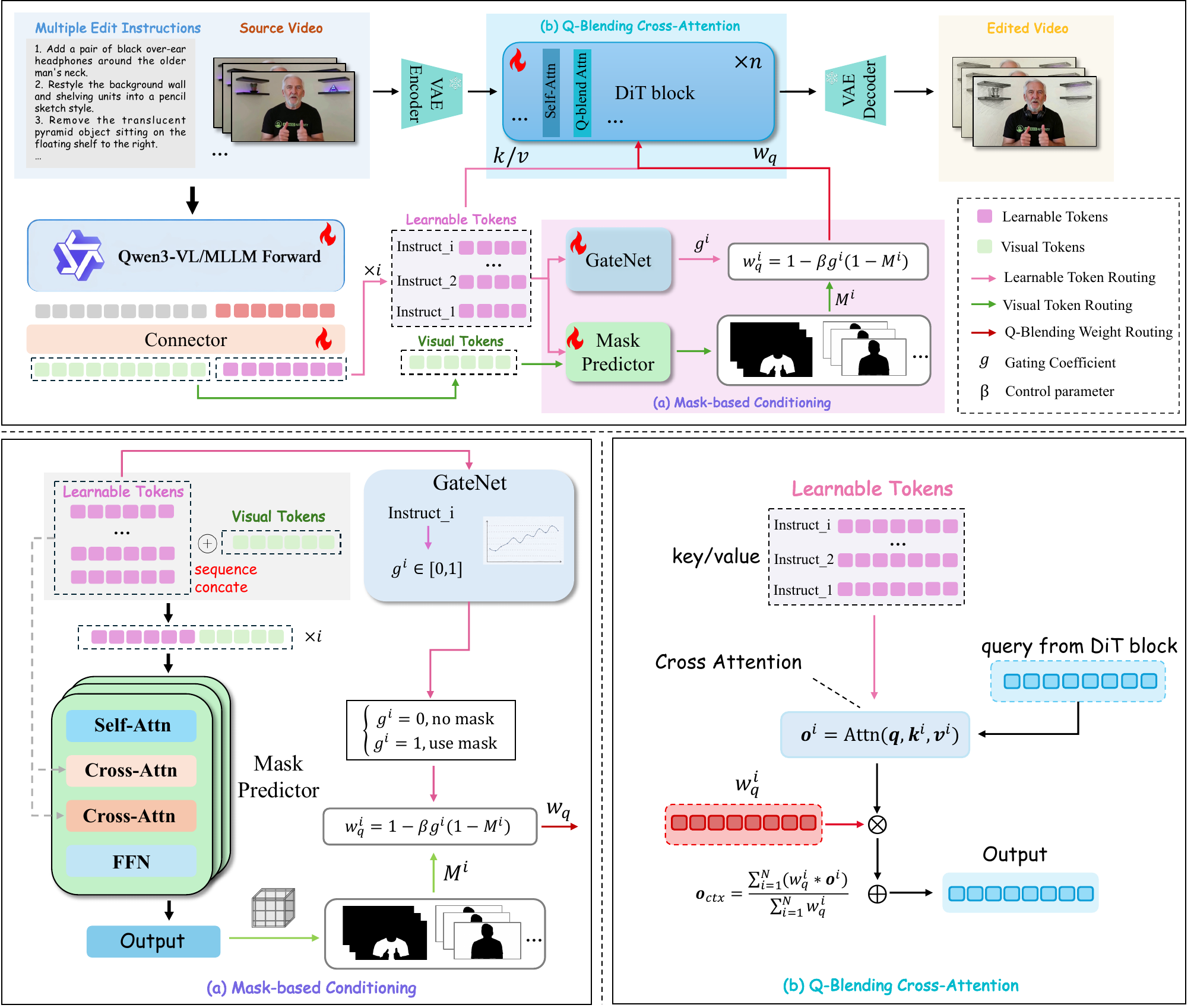}
    \vspace{0.01in}
    \caption{An overview of the proposed CoinVE-Edit framework.}
    \vspace{-0.01in}
    \label{fig:main_framework}
\end{figure}

\section{CoinVE-Edit}
\label{sec:model_architecture}

In this section, we introduce CoinVE-Edit, the newly-minted compositional instruction-guided video editing model. 
As shown in Figure~\ref{fig:main_framework}, given a source video and multiple editing instructions, the Qwen-based MLLM first jointly processes the visual content and textual instructions to extract high-level semantic representations for compositional editing. Next, a feature Connector then transforms the hidden features from the MLLM into editing-aware tokens, including instruction-related learnable tokens and visual tokens. The former serve as semantic conditions for video generation. Meanwhile, the source video is encoded into latent representations by a VAE encoder and fed into the DiT backbone for video synthesis. To enable precise multi-intent editing, the mask-based conditioning module predicts spatial-temporal editing masks, and generates routing signals to control how different instruction tokens interact with visual tokens inside the DiT blocks. Finally, the edited latent representations are decoded by the VAE decoder to produce the edited video. 
Through this pipeline, CoinVE-Edit can jointly understand multiple editing intents, associate them with corresponding video regions, and perform compositional video editing while preserving irrelevant content.

\subsection{Architecture}
\label{subsec:text_encoder}

As depicted in Figure~\ref{fig:main_framework}, CoinVE-Edit is composed of four tightly coupled modules: (i) a Qwen3-VL-based multimodal large language model (MLLM) equipped with learnable video queries, (ii) a Connector that projects the learnable-query hidden states into DiT-compatible editing-aware tokens, (iii) a conditioning Mask Predictor that predicts a spatial-temporal editing mask from the MLLM visual tokens conditioned on the visual context, and a GateNet (a lightweight instruction-level gating network) that predicts a local/global editing gate, and (iv) a DiT-based video editing backbone built upon Wan2.1-T2V-14B~\citep{wan2025wan}, in which each cross-attention block is wrapped by a Q-Blending Cross-Attention module. 

\subsubsection{Mask-based Conditioning}
\paragraph{MLLM with Learnable Video Queries.}
Given a source video $V$ and a compositional instruction $\mathcal{I}=\{I_1,\ldots,I_N\}$ of $N$ atomic edits ($N\!\ge\!2$), we adopt Qwen3-VL-8B-Instruct~\citep{qwen3vl} as our base MLLM. For the $i$-th instruction $I_i$, the input is formed as [$V$, $I_i$, $\langle\texttt{query}\rangle_{1{:}L_q}$] appended $L_q$ learnable query tokens. We read two groups of last-layer hidden states:
\begin{itemize}
\item {Visual tokens} $\mathbf{H}^i_{{vis}}$ at all video token positions, forming the size $N_v = T_v H_v W_v$;
\item {Query hidden states} $\mathbf{H}^i_{{q}}$ at the $\langle\texttt{query}\rangle$ positions, encoding the instruction jointly with the visual context.
\end{itemize}
A lightweight MLP-based Connector then projects $\mathbf{H}^i_{{q}}$ into the DiT conditioning dimension
\(
    {C}^i = \phi_{\text{con}}(\mathbf{H}^i_{{q}}).
\)
After $N$ feedforward passes of MLLM, we gather the context tensor of all $N$ instructions as $\mathbf{C}=[C^1,...,C^N]$.
The whole context tensor $\mathbf{C}$ is shared by three downstream consumers, i.e., the video DiT cross-attention, the Mask Predictor and the GateNet.

\paragraph{Mask-based Conditions.}
To prevent interference among multiple edits and to distinguish local from global/style edits, we attach two lightweight heads on top of the shared MLLM outputs: {Mask Predictor} predicts {where} the current instruction should take effect, and {GateNet} predicts {how spatially localized} that effect should be. Their outputs jointly parameterize Q-Blending (\S\ref{subsec:dit}) without altering the context $\mathbf{C}$.

Mask Predictor is a prompt-conditioned mask predictor that grounds the instruction to a spatial-temporal region on the video patch grid. For the $i$-th instruction, it takes the visual tokens $\mathbf{H}^i_{\text{vis}}$ (augmented with 3D grid positional embeddings) and the context tensor $C^i$ as key/value memory, and lets $K$ learnable mask queries interact with them through two transformer blocks (self-attention on the queries, bidirectional cross-attention with mask queries). 
The mask token is then decoded against the visual segment to yield a patch-grid mask ${M}^i\in\mathbb{R}^{T_v\times H_v\times W_v}$ (probabilities after sigmoid $\sigma(\cdot)$), where $1$ indicates the editing region. The context tensor $C^i$ conditions the prediction on the current instruction, allowing it to switch its target region when the instruction changes.

{GateNet} complements Mask Predictor by predicting whether the current edit is local (e.g., object replacement/removal, region-specific background change) or global/style (e.g., stylization, color grading, weather change), so that the mask constraint is enforced only when meaningful. The architecture is a two-layer MLP on the token-wise mean-pooled context ${C^i}$, producing a scalar gate for $i$-th instruction as follows:
\begin{equation}
    g^i = \sigma\left(\mathrm{MLP}\left(\mathrm{Mean}\left({C^i}\right)\right)\right),
\end{equation}
where $g^i\!\approx\!1$ indicates a local edit that should respect the mask, whereas $g^i\!\approx\!0$ indicates a global edit that should bypass the region constraint.
Thus, $({M}^i, g^i)$ answer ``where to edit'' and ``how strongly the where matters'', the two quantities consumed by Q-Blending.

\subsubsection{DiT-based Video Editing} 
\label{subsec:dit}
We build the video generation backbone upon Wan2.1-T2V-14B~\citep{wan2025wan}. The source video $V$ is first encoded into a latent $\mathbf{x}_0$ by the pretrained 3D VAE, and a noised latent $\mathbf{x}_t$ is drawn from the flow-matching forward process. To inject the source video as a strong structural prior, we concatenate $\mathbf{x}_0$ with $\mathbf{x}_t$ along the channel dimension before the DiT, so that layout, identity, and motion are preserved. Each DiT block contains a latent self-attention and a cross-attention that consumes the context $\mathbf{C}$ for all instructions. We wrap the latter with a Q-Blending Attention module that softly rescales its output residual per query token using $\mathbf{M}=\{{M}^i\}_{i=1}^N$ and $g=\{g^i\}_{i=1}^N$. The denoised latent is finally decoded by the VAE to produce the edited video.

\paragraph{Q-Blending Cross-Attention.}
Rather than applying hard mask biases to the attention logits, Q-Blending Cross-Attention keeps the original DiT cross-attention unchanged and modulates the attention output in a query-wise manner. For context tensor $C^i$ of each instruction, the standard cross-attention produces an instruction-specific output, which is then weighted by a blending map predicted from the mask $M^i$ and the gate weight $g^i$. These weighted outputs are summed over all $N$ instructions to form the final contextual output. This design preserves the pretrained attention distribution, avoids the instability caused by hard masking, and enables spatially adaptive composition of multiple editing instructions.

\paragraph{Mask-based Q-Blending Weight.}
Mask Predictor predicts ${M}^i$ on the MLLM patch grid $(T_v,H_v,W_v)$, whereas the DiT operates on a latent grid $(T_l,H_l,W_l)$. 
We trilinearly resample ${M}^i$ of $i$-th instruction to the latent grid and flatten it to a per-query mask. Combined with the local/global gate $g^i\in[0,1]$, we define the {Q-Blending weight} for each instruction as:
\begin{equation}
    w^i_q \;=\; 1 \,-\, \beta\, g^i\, (1 - {M}^i),\quad w_q^i\in \mathbb{R}^{T_l\times H_l\times W_l} 
    \label{eq:qblend}
\end{equation}
where $\beta\in[0,1]$ is a blending strength. Eq.~\eqref{eq:qblend} behaves as intended in three regimes: on the editing region (${M}^i\!\to\!1$) it is a no-op so the instruction takes full effect, on non-editing regions of a local edit (${M}^i\!\to\!0,\, g^i\!\to\!1$) it softly attenuates the residual to $1-\beta$ (i.e., $70\%$ under $\beta\!=\!0.3$), suppressing unintended modifications while avoiding hard zeros that would destabilize the DiT. 
For global edits ($g^i\!\to\!0$) it degenerates to $w^i_q\!\equiv\!1$, letting the instruction affect whole frame.

\paragraph{Injection into DiT Cross-Attention.}
Q-Blending is realized as a wrapper that reuses {all} pretrained projections and normalizations of the DiT original cross-attention. 
Given the block hidden states and context $\mathbf{C}=[C^1,...,C^N]$ of all $N$ instructions, we compute $\mathbf{o}^i=\mathrm{Attn}(\mathbf{q},\mathbf{k}^i,\mathbf{v}^i)$ as usual for each instruction (DiT hidden states are the query $\mathbf{q}$, $C^i$ is treated as the key $\mathbf{k}^i$ and value $\mathbf{v}^i$), then apply the obtained Q-Blending weight $w_q^i$ and sum over all $N$ instructions as:
\begin{equation}
    {\mathbf{o}}_{\text{ctx}} = \frac{\sum_{i=1}^{N}\left({w}_q^{i} \odot \mathbf{o}^{i}\right)}{\sum_{i=1}^{N} w_q^i},
\end{equation}
where $\odot$ is element-wise multiplication, and $\mathbf{o}_{\text{ctx}}$ is fed into the next transformer block. 
Note that $w_q^i$ is dynamically predicted by the Mask Predictor and GateNet for each instruction. Therefore, the Q-Blending Cross-Attention introduces no new trainable parameters into the DiT backbone.

\subsection{Training Objectives}
\label{subsec:train_obj}
Besides the standard flow-matching objective $\mathcal{L}_{\mathrm{dit}}$ for optimizing the video diffusion transformer, we introduce two auxiliary objectives to supervise the region-aware control modules, i.e., the Mask Predictor and GateNet. For the Mask Predictor, we adopt a segmentation loss composed of a binary cross-entropy loss and a Dice~\citep{milletari2016vnet} loss:
\begin{equation}
\mathcal{L}_{\mathrm{seg}} = \mathcal{L}_{\mathrm{bce}} + \mathcal{L}_{\mathrm{dice}}.
\end{equation}
This objective encourages the predicted masks to accurately localize the spatial-temporal regions corresponding to each editing instruction. For GateNet, we use a binary classification loss $\mathcal{L}_{\mathrm{gate}}$ to supervise the routing decision of whether an instruction belongs to global stylization. During training, the gradients from $\mathcal{L}_{\mathrm{seg}}$ and $\mathcal{L}_{\mathrm{gate}}$ are stopped before being propagated to the MLLM and DiT backbones, so that these auxiliary losses only update the corresponding control modules. The overall training objective is defined as:
\begin{equation}
\mathcal{L}_{\mathrm{total}} = \lambda_{\mathrm{dit}} \mathcal{L}_{\mathrm{dit}} + \lambda_{\mathrm{seg}} \mathcal{L}_{\mathrm{seg}} + \lambda_{\mathrm{gate}} \mathcal{L}_{\mathrm{gate}},
\end{equation}
where $\lambda_{\mathrm{dit}}$, $\lambda_{\mathrm{seg}}$ and $\lambda_{\mathrm{gate}}$ are the loss weight coefficients ($1.0$ for each as default).

\begin{table}[t]
\centering
\caption{
Evaluation matrix of CoinVE-Bench. The proposed metrics are grouped by evaluator type into MLLM-based checklist evaluation and specialized evaluator-based evaluation with task-specific models, covering four evaluation dimensions and eleven fine-grained metrics.}
\vspace{0.05in}
\setstretch{1.3}
\label{tab:coinve_dimension}
\resizebox{\textwidth}{!}{%
\begin{tabular}{clccl}
\hline
\textbf{Dimension} & \textbf{Metric (Abbr.)} & \textbf{Range} & \textbf{Evaluator} & \textbf{Description} \\ \hline
\multicolumn{5}{c}{\textbf{MLLM-Based (Gemini 3.6 Flash)}} \\ \hline
\multirow{3}{*}{\makecell[c]{\textbf{Editing}\\\textbf{Accuracy}\\\textbf{(per Instruction)}}} & Semantic Accuracy (SA) & {[}0,100{]} & MLLM + checklist & Correct execution of the intended edit semantics \\
 & Scope Accuracy (SPA) & {[}0,100{]} & MLLM + checklist & Correct edit localization without leakage or interference \\
 & Editing Persistence (EP) & {[}0,100{]} & MLLM + checklist & Temporal consistency of the edit throughout the video \\ \hline
\multirow{3}{*}{\makecell[c]{\textbf{Physical}\\\textbf{Naturalness}}} & Appearance Naturalness (AN) & {[}0,100{]} & MLLM + checklist & Natural blending of lighting, shadows, textures, and style \\
 & Scale Consistency (SC) & {[}0,100{]} & MLLM + checklist & Plausibility of the edited object's scale and perspective \\
 & Motion Naturalness (MN) & {[}0,100{]} & MLLM + checklist & Plausibility of motion and physical interactions \\ \hline
 \makecell[c]{\textbf{Semantic}\\\textbf{Preservation}} & Content Preservation (CP) & {[}0,100{]} & MLLM + checklist & Preservation of non-edited regions, objects, and structures \\ \hline
\multicolumn{5}{c}{\textbf{Specialized Evaluator-Based}} \\ \hline
\multirow{4}{*}{\makecell[c]{\textbf{Video}\\\textbf{Quality}}} & Aesthetic Quality (AQ) & {[}1,10{]} & Aesthetic Predictor v2.5 & Overall frame-level aesthetic appeal \\
 & Technical Quality (TQ) & (1, 100) & DOVER++ & Technical blur, noise, compression, flicker, and jitter \\
 & Comprehensive Quality (CQ) & {[}1, 5{]} & VisualQuality-R1 & Overall frame-level visual quality \\
 & Temporal Stability (TS) & {[}0, 1{]} & Optical-flow fields & Temporal motion smoothness \\ \hline
\end{tabular}
}
\vspace{-0.2in}
\end{table}

\subsection{Training Strategy}
\label{subsec:training}
To support compositional-instruction video editing, we design a progressive three-stage training paradigm that organically integrates the standalone DiT and MLLM into a unified video editing model, enabling it to perform both single-instruction and compositional-instruction edits.

\noindent
\textbf{Feature Alignment Pre-Training.}
To bridge the semantic gap between Qwen3-VL and Wan video DiT backbone, we first freeze all of the DiT weights, and only learn the learnable query tokens, connector and LoRA of MLLM.
The task is restricted to only image editing, which scales well for large-scale model training.
After this stage, the learnable tokens are adapted to the formats that can be interpreted by the cross-attention block in video DiT.

\noindent
\textbf{Single-Instruction Training.}
In the second training stage, we aim to endow the model with single-instruction video editing capability. To this end, we further unfreeze the DiT backbone and jointly optimize it with the MLLM-conditioned editing modules. Due to the limited availability of high-quality video editing data, we adopt a mixed training strategy that combines both image- and video-based single-instruction editing data. Such hybrid training enables the model to generalize well to single-instruction editing on videos, even under limited video supervision. Moreover, the resulting model weights serve as a strong prior for the next stage, facilitating the extension from single-instruction video editing to compositional-instruction video editing.

\noindent
\textbf{Compositional-Instruction Fine-Tuning.}
For the final stage, we train the model for compositional-instruction video editing, where the model is required to handle multiple editing intents. This stage involves two key capabilities: predicting the spatial regions associated with each instruction, and determining whether an instruction should trigger a global style transfer. Built upon the proposed CoinVE-200K dataset and initialized from the weights obtained in the second stage, we first freeze all model parameters and train the Mask Predictor and GateNet offline. This isolated training strategy stabilizes the learning of the auxiliary modules without disturbing the pretrained MLLM-DiT editing backbone. After offline training, the Mask Predictor achieves an IoU of $0.71$, while GateNet reaches a binary classification accuracy of $0.95$. Finally, we unfreeze the full model and conduct end-to-end fine-tuning, allowing the editing backbone, Mask Predictor, and GateNet to be jointly adapted for compositional-instruction video editing.

\section{CoinVE-Bench}
\subsection{Benchmark Construction}
To establish a unified evaluation protocol for compositional instruction-guided video editing, we construct a dedicated benchmark with $361$ high-quality source videos. The videos are collected from open-source repositories, deduplicated, and kept strictly disjoint from our training data. To ensure visual quality and editing feasibility, we retain clips with a resolution of at least $1080$p and a duration of 3--10 seconds, and remove those with severe motion blur, visible watermarks or subtitles, abrupt shot transitions, heavy occlusion, or ambiguous visual content. The resulting videos cover diverse objects, scenes, motions, and camera viewpoints. The benchmark is built around six atomic editing operations: add object, remove object, replace object, replace background, local style object, and local style background. Unlike benchmarks that focus on isolated edits, our benchmark emphasizes compositional instructions, where each instruction combines multiple atomic operations. To reduce evaluation bias, we balance the occurrence frequency of different atomic operations, the complexity of instructions with varying numbers of operations, and the diversity of operation combinations. All instructions are manually reviewed to remove ambiguous, infeasible, or visually unverifiable cases.

\subsection{Evaluation Dimensions}

As shown in Table~\ref{tab:coinve_dimension}, our evaluation matrix is structured into two complementary groups: \textbf{MLLM-based checklist evaluation} and \textbf{specialized evaluator-based evaluation}. Such design is inspired by CoVEBench~\citep{covebench}, which established the checklist-based paradigm for compositional video editing by decomposing complex instructions into fine-grained check items and judging each item with dedicated models. Following this recipe, our \textbf{MLLM-based} group evaluates the compositional correctness of instruction-guided video editing from three dimensions tailored to our task: \textbf{Editing Accuracy} (\textit{Semantic Accuracy}, \textit{Scope Accuracy}, and \textit{Editing Persistence}), \textbf{Physical Naturalness} (\textit{Appearance Naturalness}, \textit{Scale Consistency}, and \textit{Motion Naturalness}), and \textbf{Semantic Preservation} (\textit{Content Preservation}). The score of these dimensions is defined as the percentage of questions answered correctly. The \textbf{specialized evaluator-based} group, motivated by CoVEBench's use of dedicated models, complements the checklist evaluation by measuring perceptual video fidelity through the dimension of \textbf{Video Quality} (\textit{Aesthetic Quality}, \textit{Technical Quality}, \textit{Comprehensive Quality}, and \textit{Temporal Stability}). Each dimension is evaluated using a specific model, with the score scale determined by the corresponding evaluation approach. Overall, these two groups cover four dimensions and eleven fine-grained metrics, providing a unified framework for evaluating both the compositional instruction-following correctness and visual fidelity of edited videos.

\begin{table}[t]
    \centering
    \caption{Single-instruction video editing comparison on OpenVE-Bench evaluated by Gemini 2.5 Pro.}
    \vspace{0.05in}
    \label{tab:openve}
    \setlength{\tabcolsep}{1.2em}
    \resizebox{0.98\linewidth}{!}{
    \begin{tabular}{l|c|cccccc}
        \toprule
        \textbf{Model}
        & \textbf{Overall}
        & \makecell[c]{\textbf{Global}\\\textbf{Style}}
        & \makecell[c]{\textbf{Background}\\\textbf{Change}}
        & \makecell[c]{\textbf{Local}\\\textbf{Change}}
        & \makecell[c]{\textbf{Local}\\\textbf{Remove}}
        & \makecell[c]{\textbf{Local}\\\textbf{Add}}
        & \makecell[c]{\textbf{Subtitle}\\\textbf{Edit}} \\
        \midrule
        Runway       & 3.51 & 3.72          & 2.62          & 4.18          & 4.16          & 2.78          & 3.62          \\
        \midrule
        VACE         & 1.55 & 1.49          & 1.55          & 2.07          & 1.46          & 1.26          & 1.48          \\
        Ditto        & 2.25 & 4.01          & 1.68          & 2.03          & 1.53          & 1.41          & 2.81          \\
        OpenVE-Edit  & 2.57 & 3.16          & 2.36          & 2.98          & 1.85          & 2.15          & 2.91          \\
        VINO         & 2.91 & \textbf{4.05} & 1.67          & 3.12          & 3.32          & 2.66          & 2.64          \\
        OmniWeaving  & 3.01 & 3.94          & 2.03          & 3.49          & 2.93          & 2.06          & 3.58          \\
        KiwiEdit     & 3.17 & 3.60          & 2.63          & 3.87          & 3.31          & \textbf{2.83} & 2.75          \\
        SAMA         & 3.33 & 3.87          & 2.64          & \textbf{3.91} & 3.31          & 2.63          & \textbf{3.63} \\
        \midrule
        CoinVE-Edit  & \textbf{3.41} & 3.61 & \textbf{3.11} & 3.84 & \textbf{3.81} & 2.53 & 3.53 \\
        \bottomrule
    \end{tabular}
    }
    \vspace{-0.15in}
\end{table}

\section{Experiments}
\label{sec:exp}

\subsection{Implementation Details}
\label{subsec:implementation}

\textbf{Training Data.}
Our training data consists of three parts. In stage of feature alignment training, we collect open-source image editing datasets, including GPT-Image-Edit-1.5M~\citep{gpt-image-edit}, NHR-Edit~\citep{nhredit}, and Pico-Banana-400K~\citep{picobanana}, to adapt the MLLM feature space to the latent space of video DiT. GPT-Image-Edit-1.5M contains over $1.5$M source image, editing instruction, and edited image triplets constructed with advanced image generation models. NHR-Edit is a high-quality instruction-based image editing dataset mined through an automated pipeline. Pico-Banana-400K is a $400$K-scale text-guided image editing dataset built from real images with a fine-grained editing taxonomy. Since the quality of image editing data varies significantly, we filter these image editing pairs using EditScore~\citep{luo2025editscore} and keep only samples with scores higher than $8.0$, resulting in about $2$M high-quality image editing samples.
Then, in single-instructional video editing training, we collect open-source video editing data from OpenVE-3M~\citep{openve}, Ditto-1M~\citep{ditto}, and ReCo~\citep{reco}. OpenVE-3M is a large-scale instruction-following video editing dataset with about $3$M video-edit pairs, covering both spatially-aligned and non-spatially-aligned edits across multiple editing categories. Ditto-1M provides about $1$M high-fidelity synthetic video editing triplets for scaling instruction-based video editing. ReCo is a region-constrained instructional video editing dataset with over $500$K instruction-video pairs, emphasizing localized editing and preservation of non-target regions. Specifically, we use the Local Change, Background, Style, and Subtitles subsets of OpenVE-3M, the style transfer subset of Ditto-1M, and all samples from ReCo. For quality filtering, we sample frames from each video pair at $1$ FPS, compute frame-level EditScore between corresponding source-edited frame pairs, and retain video pairs with video-level scores higher than $6.0$, yielding approximately $0.9$M video editing samples.
In the final stage, we use our CoinVE-200K dataset for the compositional-instruction video editing tuning.

\noindent
\textbf{Parameter Settings.}
In CoinVE-Edit, we use the pretrained Qwen3-VL-8B-Instruct~\citep{qwen3vl} as the foundational MLLM and the Wan2.1-T2V-14B~\citep{wan2025wan} as the video DiT backbone.
In the first stage training, we only employ the image editing data, and finetune the learnable query, connector and LoRA (256 rank) in MLLM.
The learning rate is $1\times 10^{-5}$, and multi-resolution training strategy is implemented. We set the max pixel number of image as $1024\times 1024$.
The total batch size is $64$, and the optimization step is $45K$.
For the second training stage, we mix the image and video editing pairs for training, and the data sampling ratio is $1:1$.
We set the max number of frames as $49$ and the max pixels of each frame as $600\times 600$, respectively.
The LoRA in DiT is further included for optimization, and the LoRA rank in video DiT is $128$.
We set the learning rate as $1\times 10^{-5}$.
The training step is $32.5K$ with batch size of $128$.
In final compositional-instruction video editing tuning, only CoinVE-200K is exploited.
The Mask Predictor and GateNet are offline trained first to attain a satisfactory initial performance. 
Then, both modules are jointly learned with the learning rate of $5\times 10^{-6}$ and the batch size of $128$.
The model achieves stable convergence when the step is $10K$.
Experiments of all training stages are conducted on $32$ NVIDIA H200 GPUs.

\begin{figure}[t]
    \centering
    \includegraphics[width=0.99\linewidth]{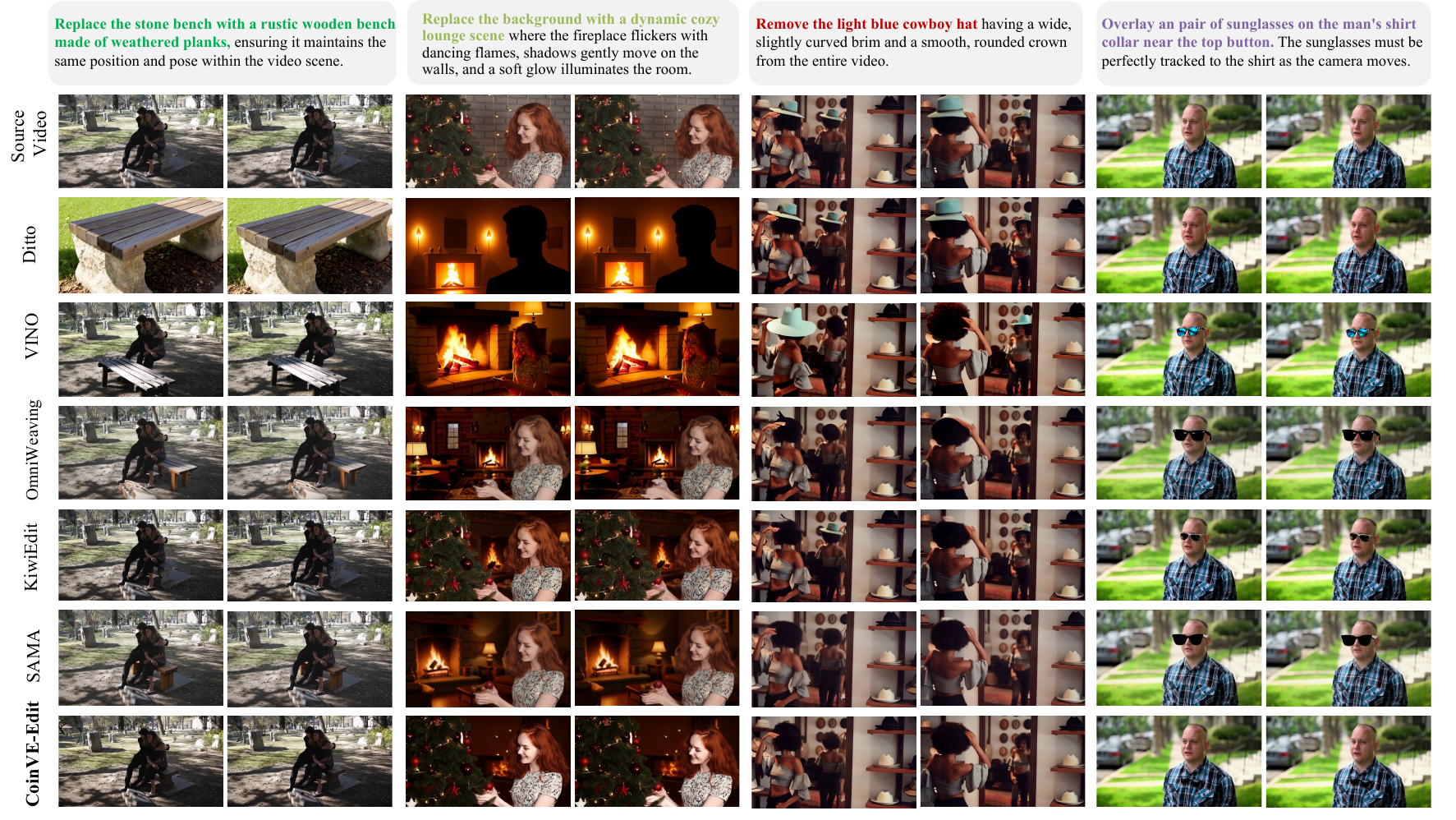}
    \vspace{0.05in}
    \caption{Quality comparison of single-instruction video editing on OpenVE-Bench.}
    \vspace{-0.15in}
    \label{fig:single_cases}
\end{figure}

\subsection{Evaluation on Single-Instruction Video Editing}
\label{subsec:eval_single}
Our CoinVE-Edit can naturally handle single instruction-guided video editing when the given instruction number $N$ equals $1$.
Here, we compare our proposal with several open-source state-of-the-art single instructional video editing methods, including VACE~\citep{jiang2025vace}, Ditto~\citep{ditto}, OpenVE-Edit~\citep{openve}, VINO~\citep{chen2026vino}, OmniWeaving~\citep{pan2026omniweaving}, KiwiEdit~\citep{lin2026kiwi} and SAMA~\citep{zhang2026sama}.
Besides, the closed-source approach Runway is also included for comparison.
Table~\ref{tab:openve} summarizes the performance of six editing tasks on OpenVE-Bench~\citep{openve}.
Overall, CoinVE-Edit attains the highest overall score $3.41$ among open-source approaches, as judged by Gemini 2.5 Pro~\citep{gemini2}.
Specifically, CoinVE-Edit demonstrates clear advantages in ``Background Change'' ($3.11$) and ``Local Remove'' ($3.81$), both of which require accurate localization of the target editing regions.
The results validate the merit of our mask-based conditioning to supply accurate region information, and the query-wise attention mechanism for precise visual content modification.

Figure~\ref{fig:single_cases} further presents four qualitative comparisons on OpenVE-Bench across different video editing models. In general, compared with existing baselines, CoinVE-Edit produces edited videos with better instruction following, higher visual quality, and more natural motion consistency.
Taking the rightmost case of ``Adding Sunglasses'' as an example, most MLLM-based SOTA methods, such as KiwiEdit and SAMA, fail to place the sunglasses at the specified region, i.e., on the shirt collar as required by the prompt. Instead, they tend to synthesize the sunglasses at the most semantically likely location and incorrectly make the man ``wear'' them. In contrast, CoinVE-Edit first localizes the target editing region and then performs content modification within the localized area. This region-aware editing manner mitigates the training-induced data bias of generative models, and enables more precise instruction-conditioned video editing.
Another advantage of CoinVE-Edit lies in its ability to handle indirect or secondary effects during video editing. With Mask Predictor, both the primary editing region and the affected regions can be identified, then Q-Blending attention injects the corresponding contextual information into the DiT for coherent synthesis. As a result, CoinVE-Edit successfully removes the ``blue cowboy hat'' not only from the woman's head but also from its mirror reflection at the corresponding location, leading to more visually plausible editing results.

\subsection{Evaluation on Compositional-Instruction Video Editing}
\label{subsec:eval_composition}

\begin{table}[t]
\centering
\caption{Compositional-Instruction Video Editing Comparisons on CoinVE-Bench.}
\vspace{0.05in}
\setstretch{1.3}
\label{tab:coinve-bench}
\resizebox{\textwidth}{!}{
\begin{tabular}{l|ccc|ccc|c|cccc}
\toprule
\multirow{2}{*}{\textbf{Model}}
    & \multicolumn{3}{c|}{\textbf{Edit. Acc.}}
    & \multicolumn{3}{c|}{\textbf{Phys. Natural.}}
    & \multicolumn{1}{c|}{\textbf{Seman. Pres.}}
    & \multicolumn{4}{c}{\textbf{Video Quality}}                            \\
\cmidrule(lr){2-4} \cmidrule(lr){5-7} \cmidrule(lr){8-8} \cmidrule(lr){9-12}
    & \textbf{SA} & \textbf{SPA} & \textbf{EP}
    & \textbf{AN} & \textbf{SC} & \textbf{MN}
    & \textbf{CP}
    & \textbf{AQ} & \textbf{TQ} & \textbf{CQ} & \textbf{TS}                  \\
\midrule
Seedance 2.0
    & 85.34          & 87.71          & 88.08
    & \textbf{93.19} & \textbf{95.84} & 92.87
    & \textbf{93.91}
    & 4.47           & 19.55 & \textbf{4.41}  & 0.62             \\
Kling O3
    & {86.91}        & 80.93          & {89.06}
    & 92.55          & 90.30          & 93.91
    & 84.51
    & \textbf{4.49}  & 18.36          & 4.37           & 0.61             \\
\midrule
VACE
    & 3.98           & 17.15          & 6.50
    & 26.69          & 13.82          & 15.21
    & 87.83
    & 4.05           & 17.59          & 4.11           & 0.62             \\
Ditto
    & 34.69          & 36.41          & 40.85
    & 35.96          & 47.79          & 38.48
    & 51.98
    & 3.59           & 17.24          & 3.96           & 0.67             \\
VINO
    & 83.63          & 66.75          & 89.06
    & 78.09          & 82.34          & 85.91
    & 61.70
    & 4.06           & 17.28          & 4.08           & 0.68             \\
OmniWeaving
    & 59.67          & 55.94          & 61.11
    & 54.49          & 66.03          & 65.10
    & 75.09
    & 3.79           & 17.95          & 3.84           & 0.62             \\
KiwiEdit
    & 76.50          & 69.92          & 80.28
    & 78.37          & 78.50          & 80.76
    & 70.31
    & 4.14           & 19.33          & 4.30           & 0.68             \\
SAMA
    & 75.58          & 73.35          & 79.63
    & 83.43          & 83.88          & 88.14
    & 90.08
    & 3.61           & 18.08          & 4.19           & 0.72    \\
\midrule
\textbf{CoinVE-Edit}
    & \textbf{87.97}          & \textbf{89.45} &  \textbf{89.60}
    & 91.85          & 91.17          & \textbf{95.30}
    & 90.83
    & 4.13           & \textbf{19.57}          & 4.31           & \textbf{0.72}      \\
\bottomrule
\end{tabular}
}
\vspace{-0.05in}
\end{table}

The core contribution of CoinVE-Edit is to handle the complex video editing task when the instruction number $N$ is larger than $1$. 
Thus, in this section, we conduct the performance comparison on the proposed CoinVE-Bench from the perspectives of \textbf{Editing Accuracy}, \textbf{Physical Naturalness}, \textbf{Semantic Preservation} and \textbf{Video Quality}.
Besides using open-source SOTA video editing approaches, we additionally include two advanced closed-source business models, i.e., Seedance 2.0~\citep{seedance2026seedance} and Kling O3~\citep{klingomini}.
For all other baselines, we concatenate the input multiple instructions into a single comma-separated prompt and feed it to the model for video editing. 
As shown in Table~\ref{tab:coinve-bench}, CoinVE-Edit outperforms all open-source approaches in terms of Editing Accuracy, Physical Naturalness and Semantic Preservation, which are highly relevant to video editing quality using compositional instruction. 
Even comparing with frontier proprietary video generation models like Seedance 2.0 and Kling O3, ours still 
manifests clear advantages in editing semantic accuracy (SA), editing scope accuracy (SPA), editing persistence (EP) and motion naturalness (MN).
In particular, CoinVE-Edit obtains $89.45$ on scope accuracy, surpassing the best competitor Seedance 2.0 by $1.74$.
The performance gain clearly confirms the effectiveness of exploring region-wise correlation across different instructions to achieve better editing accuracy. 
Moreover, video quality of CoinVE-Edit is comparable to other baselines (e.g., TQ and TS), without compromising fidelity in pursuit of higher editing accuracy.

\begin{figure}[t]
    \centering
    \includegraphics[width=0.99\linewidth]{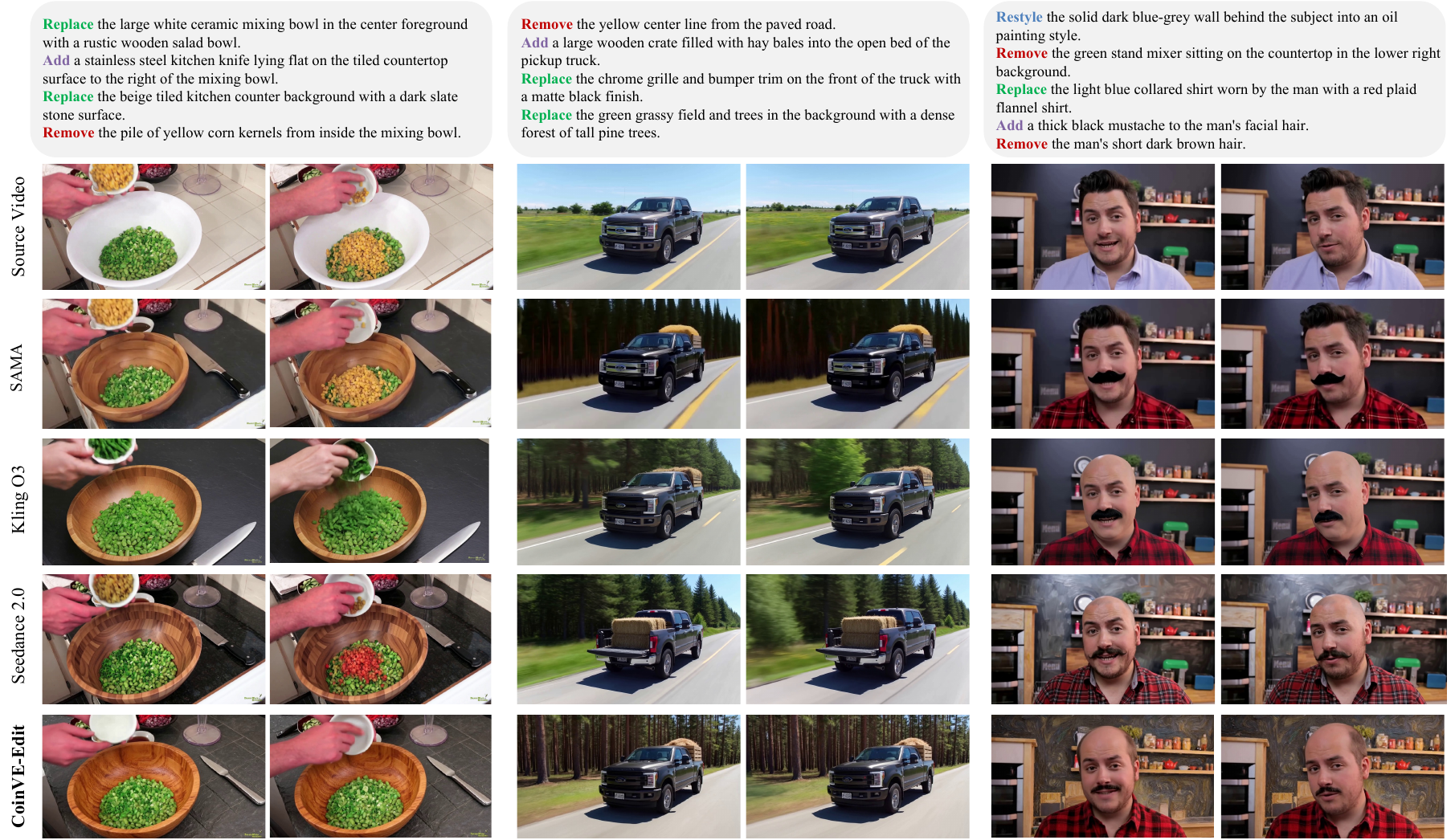}
    \vspace{0.05in}
    \caption{Quality comparison of compositional-instruction video editing on CoinVE-Bench.}
    \vspace{-0.15in}
    \label{fig:composite_cases}
\end{figure}

In Figure~\ref{fig:composite_cases}, we additionally show three visual editing examples in CoinVE-Bench across the open-source SAMA, closed-source Seedance 2.0, Kling O3 and our CoinVE-Edit.
As illustrated in the figure, both advanced Seedance 2.0 and Kling O3 cannot perfectly deal with the complex video editing scenario. 
For example, for the first case, the ``yellow corn kernels'' are not successfully removed (e.g., Kling O3 changes the color) from the bowl by the two models.
Meanwhile, some instructions could be confused with each other, leading Kling O3 to erroneously remove ``the purple glass cup'' on the top-right corner.
Instead, CoinVE-Edit precisely follows each editing instruction while preserving the consistency of non-edited regions, and even achieves better physical plausibility (e.g., Seedance 2.0 introduces unexpected ``red beans'' into the bowl, while ours prevents any contents from being poured out from the \textbf{empty} bowl).
There is also a finding that the instruction type of ``local stylization'' is easily fused with others, resulting in the ``editing leakage'' (i.e., conducting global stylization) for frontier models.
Benefiting from the CoinVE-200K for model training, ours can accurately discriminate the stylization command for local regions as shown in the third case (i.e., restyle the solid dark blue-gray wall).
We believe that our proposed dataset and method will contribute to the research community, and help raise the performance ceiling of compositional-instruction video editing, including for state-of-the-art proprietary commercial models.

\subsection{Visual Analysis and Ablations}
\label{subsec:ablation}

To better qualitatively examine our CoinVE-Edit framework for compositional instruction-guided video editing, we conduct a series of visualization and ablation studies. 

\begin{table}[t]
\centering
\caption{Performance comparison among different variants of CoinVE-Edit on CoinVE-Bench.}
\vspace{0.05in}
\setstretch{1.3}
\label{tab:ablation}
\resizebox{\textwidth}{!}{
\begin{tabular}{l|cc|ccc|ccc|c}
\toprule
\multirow{2}{*}{\textbf{Variant}}
    & \multicolumn{2}{c|}{\textbf{Components}}
    & \multicolumn{3}{c|}{\textbf{Edit. Acc.}}
    & \multicolumn{3}{c|}{\textbf{Phys. Natural.}}
    & \multicolumn{1}{c}{\textbf{Seman. Pres.}} \\
\cmidrule(lr){2-3} \cmidrule(lr){4-6} \cmidrule(lr){7-9} \cmidrule(lr){10-10}
    & \textbf{Mask} & \textbf{Q-Blending}
    & \textbf{SA} & \textbf{SPA} & \textbf{EP}
    & \textbf{AN} & \textbf{SC} & \textbf{MN}
    & \textbf{CP} \\
\midrule
Instr.-Concat.
    & $\bm{\times}$ & $\bm{\times}$
    & 82.61 & 84.43 & 89.17
    & 86.52 & 90.02 & 92.17
    & 89.30 \\
Q-Bias
    & $\bm{\checkmark}$ & $\bm{\times}$
    & 83.35 & 85.49 & 88.19
    & 87.08 & 89.64 & 91.95
    & 90.13 \\
\midrule
\textbf{CoinVE-Edit}
    & $\bm{\checkmark}$ & $\bm{\checkmark}$
    & \textbf{87.97} & \textbf{89.45} & \textbf{89.60}
    & \textbf{91.85} & \textbf{91.17} & \textbf{95.30}
    & \textbf{90.83} \\
\bottomrule
\end{tabular}
}
\end{table}

\begin{figure}[t]
    \centering
    \includegraphics[width=0.99\linewidth]{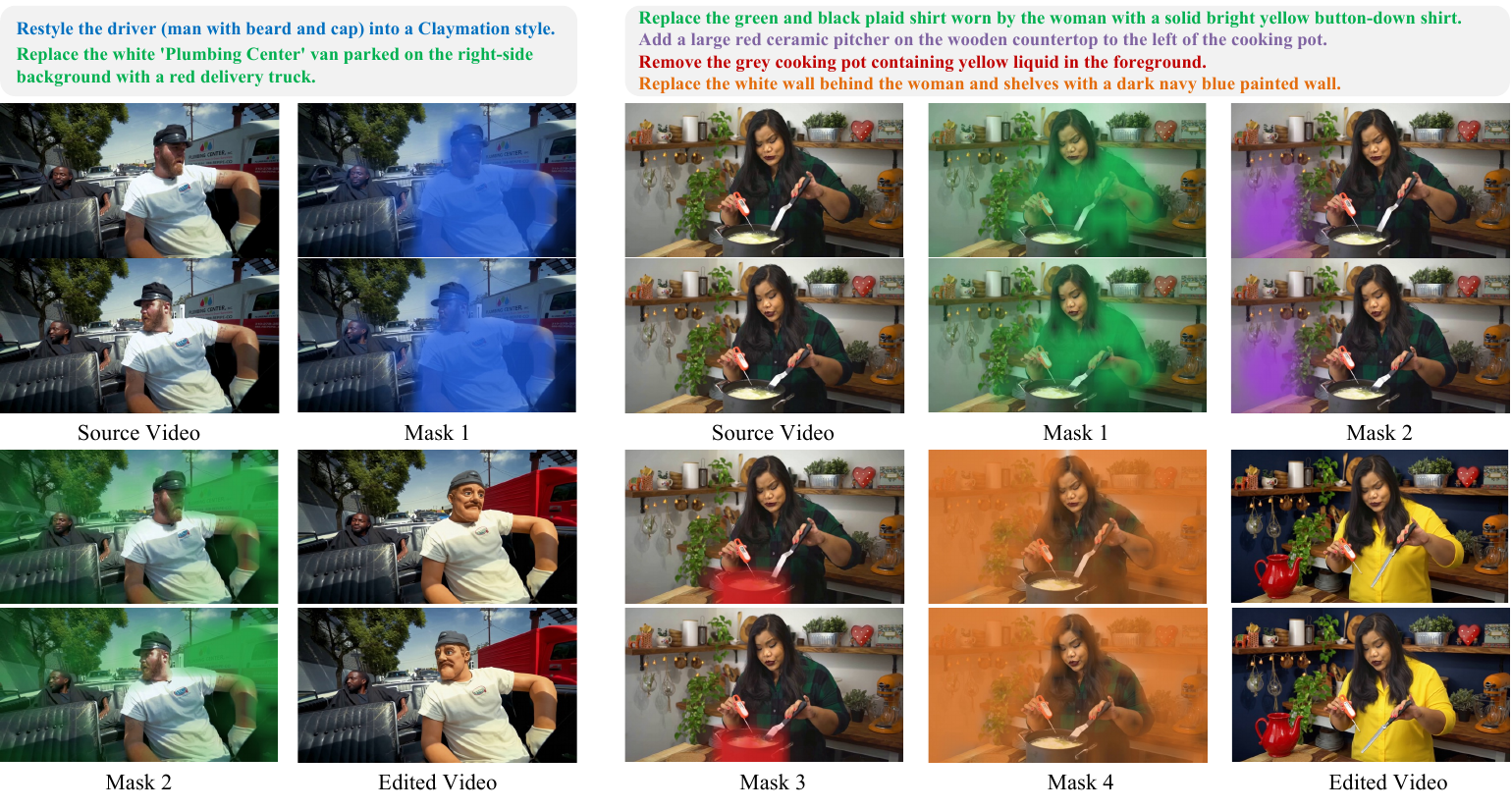}
    \caption{Visualization of predicted editing mask of each instruction for two editing examples.}
    \vspace{-0.2in}
    \label{fig:edit_mask}
\end{figure}

\textbf{Analysis of Editing Mask Prediction.}
We first visualize the predicted editing masks of two compositional-instruction video editing exemplars in Figure~\ref{fig:edit_mask}.
The estimated mask for each instruction is assigned a distinct color, which is matched with the corresponding instruction for clear association. For more faithful visualization, we upsample the masks from the latent resolution to the original frame resolution via bi-linear interpolation, and overlay them on the source video frames for display.
As shown in the visualization, CoinVE-Edit accurately localizes the target editing regions associated with different instructions. Notably, even for challenging compositional prompts containing multiple editing operations, e.g., four instructions of the second case, ours can still clearly separate the spatial regions corresponding to object addition (e.g., the red ceramic pitcher) and removal (e.g., the grey cooking pot).

\textbf{Ablation Studies.}
To investigate the efficacy of each component in CoinVE-Edit, we additionally design two baselines, i.e., Instruction-Concatenation (Instr.-Concat.) and Q-Bias.
The former merges all editing instructions into a single compound prompt and feeds it into CoinVE-Edit after removing both the mask prediction module and the Q-Blending cross-attention mechanism. 
It is trained as a standard single-instruction editing model.
The latter keeps the estimated instruction-specific masks, but replaces Q-Blending with a bias-based injection strategy, where different bias values are imposed on the masked regions corresponding to different instructions.
Table~\ref{tab:ablation} details the performances on CoinVE-Bench. 
Solely relying on single-instruction editing training (i.e., Instr.-Concat.) will face the problem of inaccurate edit localization (i.e., decreasing SPA from $89.45$ to $84.43$), as the model can only implicitly infer the target regions from contextual information through attention. 
Besides, concatenating multiple instructions into a single prompt tends to introduce interference among instruction tokens, which degrades editing accuracy (i.e., SA: $87.97 \to 82.61 $).
When involving predicted mask as edited region guidance, it is expected that Q-Bias increases the scope accuracy (SPA) for video editing.
Nevertheless, Q-Bias injects region-specific guidance through a hard mask bias in the attention layers, which may disrupt the latent feature distribution of the pretrained video backbone. 
As a consequence, it tends to degrade the physical naturalness of the edited videos, especially with respect to scale consistency (SC) and motion naturalness (MN).
By contrast, Q-Blending cross-attention serves as a soft injection that incorporates mask-aware instruction information while better preserving the original generative prior of the backbone. 
Thus, our CoinVE-Edit consistently achieves the best performance across different evaluation metrics.

\section{Conclusions}
\label{sec:conclusion}

We introduce \textbf{CoinVE-200K}, a large-scale and high-quality dataset for compositional instruction-guided video editing. CoinVE-200K contains diverse video-editing pairs with $2$ to $5$ atomic editing operations per sample, covering multiple target subjects, including humans, objects, and backgrounds, as well as edit types such as addition, removal, modification, and stylization. To ensure high-quality supervision, we design a robust data generation and filtering pipeline that emphasizes instruction faithfulness, visual quality, temporal consistency, and compositional diversity.
We also propose \textbf{CoinVE-Bench}, a dedicated benchmark for evaluating compositional video editing under multi-intent instructions. Based on CoinVE-200K, we develop \textbf{CoinVE-Edit}, a 22B compositional video editing framework that combines MLLM-based instruction understanding with DiT-based video generation. By disentangling region-aware attention across different editing instructions, CoinVE-Edit achieves precise multi-region editing while preserving irrelevant content and maintaining temporal coherence. Experiments on CoinVE-Bench demonstrate the effectiveness of our dataset, benchmark, and model, establishing a strong foundation for future research on compositional instruction-guided video editing.

\textbf{Limitations and Future Work.}
Despite its diversity, CoinVE-200K does not yet cover all possible compositional editing scenarios, such as reference-based editing, fine-grained motion editing, and complex camera control. In addition, efficient compositional editing for long videos and highly interactive scenes remains challenging. Future work will focus on expanding the editing taxonomy, improving long-range temporal consistency, reducing model cost, and developing unified video editing agents.

\bibliographystyle{abbrvnat}
\bibliography{main}
\newpage
\appendix

\end{document}